\documentclass{article}

\usepackage[dblblindworkshop, final]{neurips_2025}

\usepackage[utf8]{inputenc} 
\usepackage[T1]{fontenc}    
\usepackage{hyperref}       
\usepackage{url}            
\usepackage{booktabs}       
\usepackage{amsfonts}       
\usepackage{nicefrac}       
\usepackage{microtype}      
\usepackage{xcolor}         
\usepackage{float}
\usepackage{threeparttable}
\usepackage{colortbl}    
\usepackage{multirow}
\usepackage{graphicx}    
\usepackage{array,booktabs}
\usepackage{makecell}
\usepackage{enumitem}

\renewcommand{\rothead}[1]{\rotatebox[origin=c]{90}{\parbox{2.2cm}{\textbf{#1}}}}

\definecolor{linkcol}{HTML}{3073AD}  
\definecolor{citecol}{HTML}{3073AD} 
\definecolor{urlcol}{HTML}{3073AD} 

\definecolor{lightblue}{RGB}{32, 194, 217}
\definecolor{jku_red}{RGB}{217, 92, 76}
\definecolor{jku_blue}{RGB}{0, 132, 187}
\definecolor{jku_green}{RGB}{91, 167, 85} 
\definecolor{jku_yellow}{RGB}{241, 188, 63}
\definecolor{jku_cyan}{RGB}{79,176,191}
\definecolor{jku_grey}{RGB}{125,130,140}
\definecolor{jku_lightgreen}{RGB}{191,206,82}
\definecolor{jku_violett}{RGB}{174,97,157}

\usepackage{siunitx}
\newcommand{\belowpm}[2]{%
  \makecell[c]{$#1$\\\scriptsize $\pm\,#2$}%
}

\newcommand{\gr}{\rowcolor{jku_grey!8}}

\hypersetup{
   colorlinks=true,
   linkcolor=linkcol,
   citecolor=citecol,
   urlcolor=urlcol,
}

\usepackage[most]{tcolorbox}
\tcbset{
  colback=gray!10, colframe=gray!40,
  boxrule=0pt, arc=2pt, left=6pt,right=6pt,top=6pt,bottom=6pt
}

\newcommand{\toxleaderboard}{\url{https://huggingface.co/spaces/ml-jku/tox21_leaderboard}}

\title{Measuring AI Progress in Drug Discovery: \\ A Reproducible Leaderboard for the Tox21 Challenge}

\author{%
    \textbf{Antonia Ebner}$^{1}$ \quad
    \textbf{Christoph Bartmann}$^{1}$ \quad
    \textbf{Sonja Topf}$^{1}$ \\
    \textbf{Sohvi Luukkonen}$^{1}$ \quad
    \textbf{Johannes Schimunek}$^{1}$ \quad
    \textbf{Günter Klambauer}$^{1,2}$ \\
    \\
    $^{1}$ELLIS Unit Linz and LIT AI Lab, Institute for Machine Learning,\\
    $^{2}$Clinical Research Institute for Medical AI\\
    Johannes Kepler University, Linz, Austria\\
}

\newcommand{\stoptocwriting}{%
  \addtocontents{toc}{\protect\setcounter{tocdepth}{-5}}}
\newcommand{\resumetocwriting}{%
  \addtocontents{toc}{\protect\setcounter{tocdepth}{\arabic{tocdepth}}}}
\begin{document}

\maketitle

\begin{abstract}
    Deep learning’s rise since the early 2010s has transformed fields like computer vision and natural language processing and strongly influenced biomedical research. 
    For drug discovery specifically, a key inflection -- akin to vision’s “ImageNet moment” -- arrived in 2015, when deep neural networks surpassed traditional approaches on the Tox21 Data Challenge.
    This milestone accelerated the adoption of deep learning across the pharmaceutical industry, and today most major companies have integrated these methods into their research pipelines. 
  After the Tox21 Challenge concluded, its dataset was included in several established benchmarks, such as MoleculeNet and the Open Graph Benchmark. 
  However, during these integrations, the dataset was altered and labels were imputed or manufactured, resulting in a loss of comparability across studies. 
  Consequently, the extent to which bioactivity and toxicity prediction methods have improved over the past decade remains unclear. 
  To this end, we introduce a reproducible leaderboard\footnote{\toxleaderboard}, hosted on Hugging Face with 
  the original Tox21 Challenge dataset, together with a set of baseline and representative methods. 
  The current version of the leaderboard indicates that the original Tox21 winner -- the ensemble-based DeepTox method -- and the descriptor-based self-normalizing neural networks introduced in 2017, continue to perform competitively and rank among the top methods for toxicity prediction, leaving it unclear whether substantial progress in toxicity prediction has been achieved over the past decade.
  As part of this work, we make all baselines and evaluated models publicly accessible for inference via standardized API calls to Hugging~Face Spaces.
\end{abstract}

\stoptocwriting

\section{Introduction}
\textbf{Predicting small-molecule bioactivity, including toxicity, is a central task in drug discovery.}
Predicting these effects has been a cornerstone of computational drug discovery for decades \citep{hansch1969quantitative, goldsmith1975quantitative, hansch1987hydrophobicity}. 
Given the large number of relevant endpoints, and since determining actual efficacy and safety in the wet lab is time- and cost-intensive, accurate in silico predictions are crucial and have become key drivers of discovery \citep{ekins2004predicting, rabinowitz2008computational,mayr2018large}.
Late-stage failures remain dominated by toxicity and lack of efficacy, with other factors comparatively minor \citep{bender2021artificial}. Thus, machine learning and artificial intelligence methods should strive to develop "virtual assays", i.e., models that predict bioactivity effects as accurately as wet-lab experiments \citep{merget2017profiling,mayr2018large}.

\begin{figure}
    \centering
    \includegraphics[width=1.\linewidth]{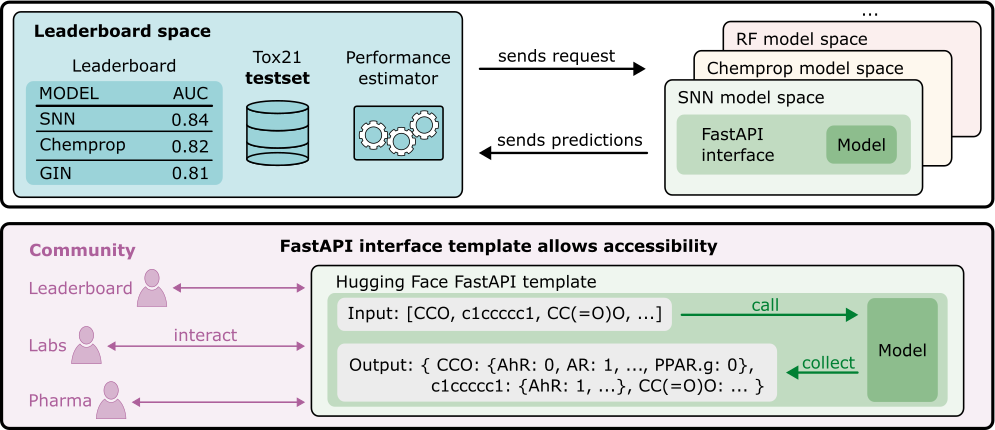}
    \caption{Overview of the Tox21 leaderboard and FastAPI interface linking model spaces with the leaderboard and external users.}
    \label{fig:overview}
\end{figure}

\textbf{The "ImageNet moment" of AI in drug discovery.} 
Somewhat akin to the "ImageNet moment" that revolutionized computer vision in 2012, deep learning emerged as the winning approach in the 2015 Tox21 Data Challenge \citep{huang2016tox21challenge}, where computational methods were tasked to predict twelve toxic effects of small molecules \citep{mayr2016deeptox}, and led to a rise of deep learning methods in drug discovery \citep{chen2018rise,walters2021critical}.
Notably, one of these toxic effects,
mitochondrial membrane disruption, could be predicted at almost experimental quality 
by the winning deep learning system, DeepTox. Since then, several large pharmaceutical companies have adopted 
deep learning and AI methods in their drug discovery pipelines \citep{chen2018rise,heyndrickx2023melloddy, valsecchi2025benchmarking, volkamer2023machine}. 

\textbf{Progress on toxicity prediction remains unclear.}
Following the Tox21 Data Challenge, the dataset has been integrated into widely used benchmarks such as MoleculeNet and DeepChem \citep{wu2018moleculenet}, and the Open Graph Benchmark \citep{hu2020open}. However, the dataset has been changed in these benchmarks in several ways: molecules have been removed, the data splits have been redesigned using different strategies, and missing labels have been replaced with zeros accompanied by a masking scheme (see Section~\ref{sec:tox21} for details).
These changes have rendered results across studies incomparable, obscuring how much progress has been achieved in molecular toxicity prediction over the last decade.

\textbf{Re-establishing a faithful evaluation setting for Tox21.}
In this work, we aim to measure the potential progress of toxicity prediction methods by comparing a set of baselines and reference methods on the original Tox21 Data Challenge dataset.
We aim at high reproducibility and sustainability, while allowing the methods maximal freedom in their software environment.
To this end, we implement the Tox21 test set as a Hugging Face leaderboard to which methods 
can be submitted. 
A method that should enter the leaderboard has to supply a model card, a reproducible training script, and -- critically -- expose an API on Hugging Face spaces that supplies predictions for API queries with small molecules coded as SMILES strings (see Figure~\ref{fig:overview}). 
In the following, we will refer to the original Tox21 
Data Challenge dataset as \texttt{Tox21-Challenge}, and
to the MoleculeNet variant as \texttt{Tox21-MoleculeNet}.

\textbf{Contributions.}
In summary, this work contributes:
\begin{itemize}
  \setlength\parskip{0pt}
  \setlength\parsep{0pt}
  \item \textbf{Re-alignment of the Tox21 benchmark.}
        We restore evaluation on the original \texttt{Tox21-Challenge} test set,
        enabling results to be compared consistently with the 2015 Data Challenge.
  \item \textbf{Analysis of benchmark drift.}
        We document differences between
        \texttt{Tox21-Challenge}, \texttt{Tox21-MoleculeNet}, and other derived variants,
        showing how dataset and metric changes have fragmented evaluation.
  \item \textbf{A reproducible, automated leaderboard.}
        We implement the first open, Hugging~Face–based evaluation pipeline that
        communicates with model APIs, executes standardized inference on the
        original test set, and stores metrics in a transparent results dataset.
  \item \textbf{Baseline re-evaluation and reference metrics.}
        We re-evaluate classic and recent baselines under the original test set and protocol,
        providing a clear picture of verifiable progress on toxicity prediction.
  \item \textbf{FastAPI template for model integration and accessibility.}
        We provide a ready-to-use FastAPI interface that enables toxicity prediction models
        to communicate with the leaderboard and facilitates external access to
        bioactivity models — for instance, by researchers and industry partners.
\end{itemize}

\section{Related Work}

\textbf{Molecular machine learning benchmarks.}
Several benchmark collections have been proposed to standardize
evaluation in molecular property prediction. 
\textsc{MoleculeNet} \citep{wu2018moleculenet} provided a unified framework
of datasets and evaluation metrics for molecular and quantum-chemical
tasks, and rapidly became the most widely used benchmark in the field.
While MoleculeNet catalyzed progress, it also modified a number of datasets, including Tox21, by changing dataset splits and preprocessing schemes. 
The \textsc{Therapeutics Data Commons (TDC)} \citep{huang2021therapeutics}
further extended this idea to a broad range of tasks across
drug discovery, drug-target interaction, and ADMET prediction \citep{van2003admet}. The \textsc{Polaris} initiative
\citep{wognum2024call} provides a benchmarking platform for
computational methods in drug discovery.
The \textsc{Open Graph Benchmark (OGB)} \citep{hu2020open}
focused on graph-structured data and enabled 
large-scale comparisons of graph neural networks.
These frameworks emphasize breadth and accessibility, but they often sacrifice historical fidelity to the original test sets or focus
on novel datasets, making comparisons across time inconsistent.

\textbf{Automated and reproducible leaderboards.}
Recent advances in open infrastructure have made benchmarking more automated and community-driven.
Platforms such as the \textsc{Open LLM Leaderboard} \citep{myrzakhan2024open} and the \textsc{LLM Eval Arena} \citep{evalarena}
enable model submissions through standardized APIs and
centralized evaluation, ensuring transparency and comparability.
Similarly, the \texttt{lm-evaluation-harness} \citep{eval-harness}
provides a reproducible interface for text-model evaluation.
Our approach builds on these principles but applies them to
molecular machine learning, integrating directly with
Hugging Face Datasets and Spaces to enable automated inference and
leaderboard updates for bioactivity models.

\textbf{Positioning of this work.}
The proposed Tox21 leaderboard combines the historical
fidelity of the original benchmark with the transparency and automation of modern leaderboards. 
It complements existing frameworks such as MoleculeNet, TDC, and OGB by providing a reproducible evaluation hub focused on toxicity prediction. 
Beyond Tox21, the infrastructure can serve as a 
blueprint for restoring evaluation consistency across
other molecular datasets where benchmark drift has occurred.

\section{The Tox21 Dataset and Its Evolution}
\label{sec:tox21}
\textbf{Overview.}
The Tox21 Data Challenge (2014--2015) was a landmark competition
for machine learning in drug discovery, as it was one of the first large, public benchmarks for toxicity prediction and catalyzed the use of deep learning in chemoinformatics \citep{huang2016modelling, mayr2016deeptox}. 
It comprised twelve in~vitro assays, related to human toxicity, spanning across the nuclear receptor (NR) and stress response (SR) pathways.
Each assay is framed as a binary classification endpoint (active vs.\ inactive) for a given target or pathway.
The dataset was compiled from the U.S.~Environmental Protection Agency (EPA), National Institutes of Health (NIH), and U.S.~Food and Drug Administration (FDA),
and contains experimentally validated toxicity measurements \citep{tox21challenge2014}.

\textbf{Dataset structure.}
The original \texttt{Tox21-Challenge} dataset contains 12,060 training compounds and 647 held-out test compounds, represented by SMILES strings and annotated for twelve endpoints:  
NR-AR, NR-AR-LBD, NR-AhR, NR-Aromatase, NR-ER, NR-ER-LBD,  
NR-PPAR-gamma, SR-ARE, SR-ATAD5, SR-HSE, SR-MMP, and SR-p53.
Not all assays were measured for every molecule, resulting in a sparse label matrix, with approximately one-third of the entries missing. 

\textbf{Evaluation.} The official challenge metric was the area under the receiver operating characteristic curve (AUC),
computed separately for each endpoint and averaged across all twelve.
The challenge organizers provided a challenging, fixed train-test split: for many test molecules, no structurally similar analogs were present in the training data, making the setup close to a cluster-based split.

\textbf{Integration into MoleculeNet and derived benchmarks.}
After the end of the Tox21-Challenge, its dataset was integrated into 
MoleculeNet and the DeepChem library \citep{wu2018moleculenet} to enable broader model benchmarking.
However, several modifications were introduced (Tables~\ref{tab:tox21_splits} and \ref{tab:tox21_all_assays}): 
\begin{enumerate}[label=(\roman*)]
    \item three new splitting strategies were introduced (random, scaffold-based, stratified), replacing the original challenge split; 
    \item the dataset was reduced from 12,060 to 8,043 or even 6,258 training molecules, while a completely new test set of 783 molecules replaced the original 647 challenge test samples; 
    \item the activity distributions in this new test set differ substantially from the original across all targets, regardless of splitting strategy; and
    \item missing labels were imputed as zeros with masking.
\end{enumerate}
Consequently, the resulting \texttt{Tox21-MoleculeNet} dataset differs from the original \texttt{Tox21-Challenge} both in data composition and in evaluation protocol, leading to results that are not directly comparable to the original leaderboard. Subsequent frameworks such as the TDC
\citep{huang2021therapeutics} and the OGB
\citep{hu2020open} adopted the \texttt{Tox21-MoleculeNet} version of Tox21. In the TDC implementation, missing labels were replaced by zeros, and the mask introduced by \cite{wu2018moleculenet} appears to be entirely removed, such that unknown assay outcomes are treated as inactive compounds. 
As a result, many reported Tox21 results are based on altered or even corrupted data.

\section{A Leaderboard for the Tox21 Challenge Dataset}

To re-align evaluation with the original \texttt{Tox21-Challenge} and
enable standardized, reproducible comparison across models, we implemented an automated leaderboard that executes evaluation directly on the original test set: \toxleaderboard.

\textbf{Design principles.}
The leaderboard design follows three guiding principles:
(i)~faithful reproduction of the original challenge protocol,
(ii)~automation and transparency of evaluation, and
(iii)~curation and approval to maintain data integrity.
All submissions are evaluated centrally on the fixed, original test set,
ensuring that performance values are directly comparable and unaffected
by local preprocessing or metric variations.

\textbf{Infrastructure and workflow.}
The leaderboard is hosted as a \texttt{Hugging~Face~Space} that communicates with user-provided model Spaces via \texttt{FastAPI}.
To request an evaluation, users submit a model entry containing
(i)~a link to their public Hugging~Face~Space exposing a \texttt{FastAPI} endpoint for inference (see Section \ref{sec:fastapi}), and
(ii)~a model card describing architecture, training data, and environment.
The request sends the SMILES strings of the 647 locally stored test compounds to the model’s API endpoint, which returns predictions to the leaderboard backend. 
The backend computes standardized metrics, and each submission is then verified by the leaderboard administrators (see Section~\ref{appsec:submission_verification}). 
Approved results are integrated into a versioned leaderboard dataset that is publicly displayed and supports filtering and sorting of models. 

\textbf{Evaluation protocol.}
Predictions are expected as floating-point values in the range~[0,1] for each of the twelve tasks.
Performance is measured by the area under the receiver operating characteristic curve (AUC), computed per endpoint and averaged across tasks, reproducing the original challenge metric.

\textbf{Tox21 training dataset release.}
To facilitate reproducible training, we provide the original \texttt{Tox21-Challenge} training data as a Hugging~Face dataset: \url{https://huggingface.co/datasets/ml-jku/tox21}.
This dataset aligns with the challenge split specification and can be used directly for model development prior to leaderboard submission.

\textbf{Baselines.}
We re-evaluate a diverse set of representative models spanning traditional machine learning, graph-based, and large language approaches.  
Our lineup includes the original \textsc{DeepTox} method (Tox21 Winner)~\citep{mayr2016deeptox}; a \textsc{Self-Normalizing Neural Network (SNN)} trained on molecular descriptors~\citep{klambauer2017self}; a graph neural network, namely a \textsc{Graph Isomorphism Network (GIN)}~\citep{xu2018powerful}, trained directly on molecular graphs; and \textsc{Chemprop}~\citep{heid2023chemprop}, which combines learned graph features from a directed message passing neural network with molecular descriptors.  
We further include classical ensemble methods such as \textsc{XGBoost}~\citep{chen2015xgboost} and \textsc{Random Forests (RF)}~\citep{breiman2001random}.
Except for DeepTox, all models listed above were trained from scratch and underwent extensive hyperparameter optimization.
We also included two pretrained models: the transformer-based \textsc{TabPFN}~\citep{hollmann2023tabpfn}, which performs inference by conditioning on a labeled support set rather than training on Tox21 data (we provide random Tox21 subsets of at most 10,000 samples per task); and a large foundation model baseline, \textsc{GPT-OSS 120B (high)}, evaluated in a zero-shot setting on SMILES representations~\citep{openai2025gptoss}.  
Together, these baselines span nearly a decade of methodological evolution in molecular toxicity prediction and lay the foundation for future community contributions and leaderboard benchmarking. See Appendix Section \ref{appsec:baselines} for model details.

\section{FastAPI Template for Model Integration and External Access}\label{sec:fastapi}
To facilitate seamless integration of new toxicity prediction models, we provide
a minimal \texttt{FastAPI} template that standardizes communication between models
and the leaderboard backend. Each model exposes a single \texttt{/predict} endpoint
that accepts a list of SMILES strings and returns a nested JSON dictionary of
predictions for the twelve Tox21 targets. The template is lightweight and modular:
developers only need to adapt the \texttt{predict\_fn} function with model-specific
preprocessing and inference code, while the input--output structure remains fixed.
This design ensures compatibility with the leaderboard orchestrator and enables
uniform access to different model architectures.

Beyond leaderboard evaluation, the \texttt{FastAPI} interface also facilitates
external access to toxicity prediction models by other clients, such as
pharmaceutical partners or research groups, without requiring them to install or
configure the underlying model code. Each provided model space serves as a working example that can be cloned to deploy new models with minimal modification; e.g., see \url{https://huggingface.co/spaces/ml-jku/tox21_gin_classifier}.
Together, these templates form a reusable interface layer that supports both
reproducible benchmarking and real-world integration of molecular prediction models.


\begin{table}[h]
\centering
\resizebox{\linewidth}{!}{%
\begin{threeparttable}
\footnotesize
\setlength{\tabcolsep}{2pt}
\renewcommand{\arraystretch}{1.15}
\caption{\textbf{Leaderboard results on the Tox21 test set (AUC $\uparrow$).} For each target, we report the mean ± standard deviation over 5 runs with different random seeds. The overall score is the median across reruns for the averaged target AUCs with the median absolute deviation (MAD). Note that minor differences between performance values in this table and the leaderboard on HuggingFace might occur because a) here we report the average over five runs whereas the leaderboard uses the median model among those five and b) the models are run on different computing environments. 
}
\label{tab:main_results}

\begin{tabular}{l c| *{12}{c}}
\toprule
\textbf{Task} & \rothead{Median $\pm$ MAD} & \rothead{NR-AR} & \rothead{NR-AR-LBD} & \rothead{NR-AhR} & \rothead{NR-Aromatase} &\rothead{NR-ER} & \rothead{NR-ER-LBD} & \rothead{NR-PPAR-gamma} & 
\rothead{SR-ARE} &  \rothead{SR-ATAD5} & 
\rothead{SR-HSE} & \rothead{SR-MMP} & \rothead{SR-p53} \\
\midrule
Tox21 Winner$^1$
& .846
  & .807
  & .879
  & .928
  & .834
  & .810
  & .814
  & .861
  & .840
  & .793
  & .865
  & .942
  & .862
  \\
\gr SNN$^2$
& \belowpm{.844}{.002}
  & \belowpm{.854}{.013}
  & \belowpm{.933}{.011}
  & \belowpm{.896}{.005}
  & \belowpm{.786}{.012}
  & \belowpm{.795}{.013}
  & \belowpm{.803}{.007}
  & \belowpm{.840}{.010}
  & \belowpm{.794}{.007}
  & \belowpm{.813}{.002}
  & \belowpm{.825}{.007}
  & \belowpm{.941}{.004}
  & \belowpm{.856}{.008}\\

RF$^3$ 
& \belowpm{.829}{.001}
  & \belowpm{.783}{.007}
  & \belowpm{.768}{.016}
  & \belowpm{.917}{.002}
  & \belowpm{.820}{.003}
  & \belowpm{.814}{.001}
  & \belowpm{.774}{.003}
  & \belowpm{.827}{.007}
  & \belowpm{.800}{.002}
  & \belowpm{.811}{.006}
  & \belowpm{.842}{.002}
  & \belowpm{.946}{.001}
  & \belowpm{.851}{.002}\\
 
\gr XGBoost$^4$
& \belowpm{.823}{.010}
  & \belowpm{.693}{.046}
  & \belowpm{.824}{.038}
  & \belowpm{.912}{.004}
  & \belowpm{.808}{.016}
  & \belowpm{.810}{.006}
  & \belowpm{.788}{.011}
  & \belowpm{.813}{.032}
  & \belowpm{.811}{.004}
  & \belowpm{.811}{.009}
  & \belowpm{.813}{.021}
  & \belowpm{.944}{.002}
  & \belowpm{.821}{.015}\\

Chemprop$^5$
& \belowpm{.815}{.005}
  & \belowpm{.894}{.006}
  & \belowpm{.846}{.016}
  & \belowpm{.873}{.013}
  & \belowpm{.780}{.015}
  & \belowpm{.805}{.015}
  & \belowpm{.754}{.022}
  & \belowpm{.793}{.028}
  & \belowpm{.754}{.005}
  & \belowpm{.783}{.007}
  & \belowpm{.792}{.031}
  & \belowpm{.926}{.010}
  & \belowpm{.818}{.016}\\
  
\gr GIN$^6$
& \belowpm{.811}{.003}
  & \belowpm{.892}{.007}
  & \belowpm{.805}{.009}
  & \belowpm{.906}{.018}
  & \belowpm{.795}{.011}
  & \belowpm{.773}{.010}
  & \belowpm{.757}{.028}
  & \belowpm{.768}{.017}
  & \belowpm{.745}{.007}
  & \belowpm{.800}{.016}
  & \belowpm{.765}{.025}
  & \belowpm{.925}{.006}
  & \belowpm{.806}{.018}\\
 TabPFN$^7$
& \belowpm{.807}{.005}
  & \belowpm{.746}{.034}
  & \belowpm{.703}{.034}
  & \belowpm{.898}{.003}
  & \belowpm{.785}{.020}
  & \belowpm{.783}{.018}
  & \belowpm{.787}{.017}
  & \belowpm{.790}{.011}
  & \belowpm{.797}{.007}
  & \belowpm{.790}{.012}
  & \belowpm{.837}{.019}
  & \belowpm{.948}{.005}
  & \belowpm{.826}{.016}\\
  
\gr GPT-OSS$^8$
  & \belowpm{.703}{.033}
  & \belowpm{.625}{.042}
  & \belowpm{.673}{.042}
  & \belowpm{.829}{0.008}
  & \belowpm{.715}{.011}
  & \belowpm{.656}{.008}
  & \belowpm{.705}{.018}
  & \belowpm{.659}{.031}
  & \belowpm{.701}{.010}
  & \belowpm{.667}{.009}
  & \belowpm{.728}{.039}
  & \belowpm{.765}{.010}
  & \belowpm{.710}{.011}
 
   \\

\bottomrule
\end{tabular}

\footnotesize
$^1$ taken from \citet{mayr2016deeptox}, 
$^2$ reimplemented from \citet{klambauer2017self}, 
$^3$ \citet{breiman2001random}
$^4$ \citet{chen2015xgboost},
$^5$ \citet{heid2023chemprop}, 
$^6$ \citet{xu2018powerful}, 
$^7$ \citet{hollmann2023tabpfn},
$^8$ GPT-OSS 120B high (zero-shot) \citep{openai2025gptoss}
\end{threeparttable}
} 
\end{table}

\section{Results}
\label{sec:results}

Table~\ref{tab:main_results} summarizes the performance of our baseline toxicity prediction models on the original \texttt{Tox21-Challenge} test set.  
Overall, descriptor-based architectures remain highly competitive nearly a decade after the original challenge. 
Comparing the implemented baselines with the Tox21 winner method DeepTox, the SNN achieves remarkably strong performance despite consisting of a single model, whereas DeepTox represents a large ensemble of networks.  
RFs follow closely and even outperform XGBoost, which is noteworthy since gradient-boosted trees typically dominate in tabular domains, a reversal of the usual trend that has also been reported in a previous Tox21 study~\citep{wu2021trade}.
The transformer-based TabPFN model reaches competitive results without task-specific training, demonstrating the potential of pre-trained tabular transformers and in-context learning for molecular prediction tasks.  
Finally, the zero-shot GPT-OSS 120B model achieves reasonably good results considering the complete absence of fine-tuning or explicit molecular supervision, suggesting that large language models encode a degree of chemical regularity even when trained primarily on textual data.

The results in Table~\ref{tab:main_results} differ slightly from those on the Hugging Face leaderboard. Table~\ref{tab:main_results} reports the median performance across five independent runs conducted on our local infrastructure, whereas the leaderboard shows the score from a single run of the median-performing model, evaluated via the standardized Hugging Face pipeline. The observed discrepancies are attributable to hardware- and system-level sources of nondeterminism.

\section{Conclusion and Outlook}

The presented leaderboard restores faithful evaluation of the original
\texttt{Tox21-Challenge} and enables reproducible comparison of
bioactivity models under identical conditions.
By combining historical fidelity with automated benchmarking on
Hugging~Face Spaces, it addresses long-standing inconsistencies in how
molecular toxicity models have been evaluated.
Our findings indicate that it remains unclear how much genuine progress
has been achieved over the past decade, 
as a method proposed in
2014 still perform competitively on the original benchmark.
This observation suggests that similar re-evaluations may be warranted
for other bioactivity prediction endpoints such as SIDER or MUV,
where benchmark drift may likewise obscure scientific progress.
Beyond serving as a reliable reference for supervised toxicity
prediction, the framework can be readily extended to few-shot and
zero-shot evaluation settings, as illustrated by our pre-trained
baselines.
Such extensions will allow systematic assessment of foundation models and in-context learning approaches for molecular prediction \citep{chenmeta, schimunek2023contextenriched, schimunek2025mhnfs, stanley2021fs}, contributing to a larger landscape of
bioactivity model evaluation.

\section*{Acknowledgements}
The ELLIS Unit Linz, the LIT AI Lab, and the Institute for Machine Learning are supported by the Federal State of Upper Austria. We thank the projects FWF AIRI FG 9-N (10.55776/FG9), AI4GreenHeatingGrids (FFG- 899943), Stars4Waters (HORIZON-CL6-2021-CLIMATE-01-01), and FWF Bilateral Artificial Intelligence (10.55776/COE12). We thank NXAI GmbH, Audi AG, Silicon Austria Labs (SAL), Merck Healthcare KGaA, GLS (Univ. Waterloo), T\"{U}V Holding GmbH, Software Competence Center Hagenberg GmbH, dSPACE GmbH, TRUMPF SE + Co. KG.

\clearpage
\bibliography{bib}

@article{huang2016tox21challenge,
  title={Tox21Challenge to build predictive models of nuclear receptor and stress response pathways as mediated by exposure to environmental chemicals and drugs},
  author={Huang, Ruili and Xia, Menghang and Nguyen, Dac-Trung and Zhao, Tongan and Sakamuru, Srilatha and Zhao, Jinghua and Shahane, Sampada A and Rossoshek, Anna and Simeonov, Anton},
  journal={Frontiers in Environmental Science},
  volume={3},
  pages={85},
  year={2016},
  publisher={Frontiers Media SA}
}

@article{wognum2024call,
  title={A call for an industry-led initiative to critically assess machine learning for real-world drug discovery},
  author={Wognum, Cas and Ash, Jeremy R and Aldeghi, Matteo and Rodr{\'\i}guez-P{\'e}rez, Raquel and Fang, Cheng and Cheng, Alan C and Price, Daniel J and Clevert, Djork-Arn{\'e} and Engkvist, Ola and Walters, W Patrick},
  journal={Nature Machine Intelligence},
  volume={6},
  number={10},
  pages={1120--1121},
  year={2024},
  publisher={Nature Publishing Group UK London}
}

@article{wu2021trade,
  title        = {Trade-off predictivity and explainability for machine-learning powered predictive toxicology: An in-depth investigation with Tox21 data sets},
  author       = {Wu, Leihong and Huang, Ruili and Tetko, Igor V and Xia, Zhonghua and Xu, Joshua and Tong, Weida},
  year         = 2021,
  journal      = {Chemical Research in Toxicology},
  publisher    = {ACS Publications},
  volume       = 34,
  number       = 2,
  pages        = {541--549}
}

@inproceedings{
hollmann2023tabpfn,
title={Tab{PFN}: A Transformer That Solves Small Tabular Classification Problems in a Second},
author={Noah Hollmann and Samuel M{\"u}ller and Katharina Eggensperger and Frank Hutter},
booktitle={The Eleventh International Conference on Learning Representations },
year={2023},
url={https://openreview.net/forum?id=cp5PvcI6w8_}
}

@inproceedings{stanley2021fs,
  title        = {Fs-mol: A few-shot learning dataset of molecules},
  author       = {Stanley, Megan and Bronskill, John F and Maziarz, Krzysztof and Misztela, Hubert and Lanini, Jessica and Segler, Marwin and Schneider, Nadine and Brockschmidt, Marc},
  year         = 2021,
  booktitle    = {Thirty-fifth Conference on Neural Information Processing Systems Datasets and Benchmarks Track (Round 2)}
}

@article{merget2017profiling,
  title={Profiling prediction of kinase inhibitors: toward the virtual assay},
  author={Merget, Benjamin and Turk, Samo and Eid, Sameh and Rippmann, Friedrich and Fulle, Simone},
  journal={Journal of medicinal chemistry},
  volume={60},
  number={1},
  pages={474--485},
  year={2017},
  publisher={ACS Publications}
}

@article{walters2021critical,
  title={Critical assessment of AI in drug discovery},
  author={Walters, W Patrick and Barzilay, Regina},
  journal={Expert opinion on drug discovery},
  volume={16},
  number={9},
  pages={937--947},
  year={2021},
  publisher={Taylor \& Francis}
}

@article{bender2021artificial,
  title        = {Artificial intelligence in drug discovery: what is realistic, what are illusions? Part 1: ways to make an impact, and why we are not there yet},
  author       = {Bender, Andreas and Cort{\'e}s-Ciriano, Isidro},
  year         = 2021,
  journal      = {Drug Discovery Today},
  publisher    = {Elsevier},
  volume       = 26,
  number       = 2,
  pages        = {511--524}
}

@article{hansch1969quantitative,
  title        = {Quantitative approach to biochemical structure-activity relationships},
  author       = {Hansch, Corwin},
  year         = 1969,
  journal      = {Accounts of chemical research},
  publisher    = {ACS Publications},
  volume       = 2,
  number       = 8,
  pages        = {232--239}
}

@article{chen2018rise,
  title        = {The rise of deep learning in drug discovery},
  author       = {Chen, Hongming and Engkvist, Ola and Wang, Yinhai and Olivecrona, Marcus and Blaschke, Thomas},
  year         = 2018,
  journal      = {Drug Discovery Today},
  publisher    = {Elsevier},
  volume       = 23,
  number       = 6,
  pages        = {1241--1250}
}

@article{mayr2016deeptox,
  title        = {DeepTox: toxicity prediction using deep learning},
  author       = {Mayr, Andreas and Klambauer, G{\"u}nter and Unterthiner, Thomas and Hochreiter, Sepp},
  year         = 2016,
  journal      = {Frontiers in Environmental Science},
  publisher    = {Frontiers},
  volume       = 3,
  pages        = 80,
  doix          = {10.3389/fenvs.2015.00080}
}

@inproceedings{klambauer2017self,
  title        = {Self-Normalizing Neural Networks},
  author       = {Klambauer, G{\"u}nter and Unterthiner, Thomas and Mayr, Andreas and Hochreiter, Sepp},
  year         = 2017,
  booktitle    = {Advances in Neural Information Processing Systems 30},
  pages        = {972--981}
}

@article{mayr2018large,
  title        = {Large-scale comparison of machine learning methods for drug target prediction on {ChEMBL}},
  author       = {Mayr, Andreas and Klambauer, G{\"u}nter and Unterthiner, Thomas and Steijaert, Marvin and Wegner, J{\"o}rg K and Ceulemans, Hugo and Clevert, Djork-Arn{\'e} and Hochreiter, Sepp},
  year         = 2018,
  journal      = {Chemical Science},
  publisher    = {Royal Society of Chemistry},
  volume       = 9,
  number       = 24,
  pages        = {5441--5451},
  doix          = {10.1039/c8sc00148k}
}

@article{schimunek2023contextenriched,
  title        = {Context-enriched molecule representations improve few-shot drug discovery},
  author       = {Johannes Schimunek and Philipp Seidl and Lukas Friedrich and Daniel Kuhn and Friedrich Rippmann and Sepp Hochreiter and G{\"u}nter Klambauer},
  year         = 2023,
  journal      = {International Conference on Learning Representations}
}

@article{wu2018moleculenet,
  title        = {MoleculeNet: a benchmark for molecular machine learning},
  author       = {Wu, Zhenqin and Ramsundar, Bharath and Feinberg, Evan N and Gomes, Joseph and Geniesse, Caleb and Pappu, Aneesh S and Leswing, Karl and Pande, Vijay},
  year         = 2018,
  journal      = {Chemical Science},
  publisher    = {Royal Society of Chemistry},
  volume       = 9,
  number       = 2,
  pages        = {513--530}
}

@article{huang2021therapeutics,
  title={Therapeutics data commons: Machine learning datasets and tasks for drug discovery and development},
  author={Huang, Kexin and Fu, Tianfan and Gao, Wenhao and Zhao, Yue and Roohani, Yusuf and Leskovec, Jure and Coley, Connor W and Xiao, Cao and Sun, Jimeng and Zitnik, Marinka},
  journal={arXiv preprint arXiv:2102.09548},
  year={2021}
}

@article{hansch1987hydrophobicity,
  title={Hydrophobicity and central nervous system agents: on the principle of minimal hydrophobicity in drug design},
  author={Hansch, Corwin and Bj{\"o}rkroth, JP and Leo, A},
  journal={Journal of pharmaceutical sciences},
  volume={76},
  number={9},
  pages={663--687},
  year={1987},
  publisher={Wiley Online Library}
}

@article{goldsmith1975quantitative,
  title={Quantitative prediction of drug toxicity in humans from toxicology in small and large animals},
  author={Goldsmith, Michael A and Slavik, Milan and Carter, Stephen K},
  journal={Cancer Research},
  volume={35},
  number={5},
  pages={1354--1364},
  year={1975},
  publisher={The American Association for Cancer Research}
}

@article{rabinowitz2008computational,
  title={Computational molecular modeling for evaluating the toxicity of environmental chemicals: prioritizing bioassay requirements},
  author={Rabinowitz, James R and Goldsmith, Michael-Rock and Little, Stephen B and Pasquinelli, Melissa A},
  journal={Environmental Health Perspectives},
  volume={116},
  number={5},
  pages={573--577},
  year={2008},
  publisher={National Institute of Environmental Health Sciences}
}

@article{ekins2004predicting,
  title={Predicting undesirable drug interactions with promiscuous proteins in silico},
  author={Ekins, Sean},
  journal={Drug discovery today},
  volume={9},
  number={6},
  pages={276--285},
  year={2004},
  publisher={Elsevier}
}

@article{volkamer2023machine,
  title={Machine learning for small molecule drug discovery in academia and industry},
  author={Volkamer, Andrea and Riniker, Sereina and Nittinger, Eva and Lanini, Jessica and Grisoni, Francesca and Evertsson, Emma and Rodr{\'\i}guez-P{\'e}rez, Raquel and Schneider, Nadine},
  journal={Artificial Intelligence in the Life Sciences},
  volume={3},
  pages={100056},
  year={2023},
  publisher={Elsevier}
}

@article{hu2020open,
  title={Open graph benchmark: Datasets for machine learning on graphs},
  author={Hu, Weihua and Fey, Matthias and Zitnik, Marinka and Dong, Yuxiao and Ren, Hongyu and Liu, Bowen and Catasta, Michele and Leskovec, Jure},
  journal={Advances in neural information processing systems},
  volume={33},
  pages={22118--22133},
  year={2020}
}

@article{myrzakhan2024open,
  title={Open-llm-leaderboard: From multi-choice to open-style questions for llms evaluation, benchmark, and arena},
  author={Myrzakhan, Aidar and Bsharat, Sondos Mahmoud and Shen, Zhiqiang},
  journal={arXiv preprint arXiv:2406.07545},
  year={2024}
}

@misc{evalarena,
  title = {{E}val-{A}rena: noise and errors on LLM evaluations},
  author = {Sida I. Wang and Alex Gu and Lovish Madaan and Dieuwke Hupkes and Jiawei Liu and Yuxiang Wei and Naman Jain and Yuhang Lai and Sten Sootla and Ofir Press and Baptiste Rozière and Gabriel Synnaeve},
  year = {2024},
  publisher = {GitHub},
  journal = {GitHub repository},
  howpublished = {\url{https://github.com/crux-eval/eval-arena}}
}

@misc{eval-harness,
  author       = {Gao, Leo and Tow, Jonathan and Abbasi, Baber and Biderman, Stella and Black, Sid and DiPofi, Anthony and Foster, Charles and Golding, Laurence and Hsu, Jeffrey and Le Noac'h, Alain and Li, Haonan and McDonell, Kyle and Muennighoff, Niklas and Ociepa, Chris and Phang, Jason and Reynolds, Laria and Schoelkopf, Hailey and Skowron, Aviya and Sutawika, Lintang and Tang, Eric and Thite, Anish and Wang, Ben and Wang, Kevin and Zou, Andy},
  title        = {The Language Model Evaluation Harness},
  month        = 07,
  year         = 2024,
  publisher    = {Zenodo},
  version      = {v0.4.3},
  doi          = {10.5281/zenodo.12608602},
  url          = {https://zenodo.org/records/12608602}
}

@article{huang2016modelling,
  title={Modelling the Tox21 10 K chemical profiles for in vivo toxicity prediction and mechanism characterization},
  author={Huang, Ruili and Xia, Menghang and Sakamuru, Srilatha and Zhao, Jinghua and Shahane, Sampada A and Attene-Ramos, Matias and Zhao, Tongan and Austin, Christopher P and Simeonov, Anton},
  journal={Nature communications},
  volume={7},
  number={1},
  pages={10425},
  year={2016},
  publisher={Nature Publishing Group UK London}
}

@misc{tox21challenge2014,
  title        = {{Tox21 Data Challenge}},
  author       = {{U.S. National Institutes of Health (NIH)} and {Environmental Protection Agency (EPA)} and {Food and Drug Administration (FDA)}},
  year         = {2014},
  howpublished = {\url{https://tripod.nih.gov/tox21/challenge/}},
  note         = {Accessed: 2025-10-07},
}

@misc{openai2025gptoss120bgptoss20bmodel,
      title={gpt-oss-120b \& gpt-oss-20b Model Card}, 
      author={OpenAI},
      year={2025},
      eprint={2508.10925},
      archivePrefix={arXiv},
      primaryClass={cs.CL},
      url={https://arxiv.org/abs/2508.10925}, 
}

@inproceedings{kwon2023vllm,
  title={Efficient Memory Management for Large Language Model Serving with PagedAttention},
  author={Woosuk Kwon and Zhuohan Li and Siyuan Zhuang and Ying Sheng and Lianmin Zheng and Cody Hao Yu and Joseph E. Gonzalez and Hao Zhang and Ion Stoica},
  booktitle={Proceedings of the ACM SIGOPS 29th Symposium on Operating Systems Principles},
  year={2023}
}

@article{heid2023chemprop,
  title={Chemprop: a machine learning package for chemical property prediction},
  author={Heid, Esther and Greenman, Kevin P and Chung, Yunsie and Li, Shih-Cheng and Graff, David E and Vermeire, Florence H and Wu, Haoyang and Green, William H and McGill, Charles J},
  journal={Journal of Chemical Information and Modeling},
  volume={64},
  number={1},
  pages={9--17},
  year={2023},
  publisher={ACS Publications}
}

@article{chen2015xgboost,
  title={Xgboost: extreme gradient boosting},
  author={Chen, Tianqi and He, Tong and Benesty, Michael and Khotilovich, Vadim and Tang, Yuan and Cho, Hyunsu and Chen, Kailong and Mitchell, Rory and Cano, Ignacio and Zhou, Tianyi and others},
  journal={R package version 0.4-2},
  volume={1},
  number={4},
  pages={1--4},
  year={2015}
}

@article{breiman2001random,
  title={Random forests},
  author={Breiman, Leo},
  journal={Machine learning},
  volume={45},
  number={1},
  pages={5--32},
  year={2001},
  publisher={Springer}
}

@article{xu2018powerful,
  title={How powerful are graph neural networks?},
  author={Xu, Keyulu and Hu, Weihua and Leskovec, Jure and Jegelka, Stefanie},
  journal={arXiv preprint arXiv:1810.00826},
  year={2018}
}

@misc{openai2025gptoss,	
title = {{GPT-OSS}: {Open-Weight} {Model} {Release} ({GPT-OSS-20B}, {GPT-OSS-120B})},
  author = {OpenAI},
  year = {2025},
  month = aug,
  howpublished = {Model documentation},
  url = {https://openai.com/index/introducing-gpt-oss/}	
}

@article{valsecchi2025benchmarking,
  title={Benchmarking molecular conformer augmentation with context-enriched training: graph-based transformer versus GNN models},
  author={Valsecchi, Cecile and Arjona-Medina, Jose A and Dyubankova, Natalia and Nugmanov, Ramil},
  journal={Journal of Cheminformatics},
  volume={17},
  number={1},
  pages={80},
  year={2025},
  publisher={Springer}
}

@article{heyndrickx2023melloddy,
  title={Melloddy: Cross-pharma federated learning at unprecedented scale unlocks benefits in qsar without compromising proprietary information},
  author={Heyndrickx, Wouter and Mervin, Lewis and Morawietz, Tobias and Sturm, No{\'e} and Friedrich, Lukas and Zalewski, Adam and Pentina, Anastasia and Humbeck, Lina and Oldenhof, Martijn and Niwayama, Ritsuya and others},
  journal={Journal of chemical information and modeling},
  volume={64},
  number={7},
  pages={2331--2344},
  year={2023},
  publisher={ACS Publications}
}

@article{van2003admet,
  title={ADMET in silico modelling: towards prediction paradise?},
  author={Van De Waterbeemd, Han and Gifford, Eric},
  journal={Nature reviews Drug discovery},
  volume={2},
  number={3},
  pages={192--204},
  year={2003},
  publisher={Nature Publishing Group UK London}
}

@inproceedings{mitchell2019model,
  title={Model cards for model reporting},
  author={Mitchell, Margaret and Wu, Simone and Zaldivar, Andrew and Barnes, Parker and Vasserman, Lucy and Hutchinson, Ben and Spitzer, Elena and Raji, Inioluwa Deborah and Gebru, Timnit},
  booktitle={Proceedings of the conference on fairness, accountability, and transparency},
  pages={220--229},
  year={2019}
}

@book{Ramsundar-et-al-2019,
    title={Deep Learning for the Life Sciences},
    author={Bharath Ramsundar and Peter Eastman and Patrick Walters and Vijay Pande and Karl Leswing and Zhenqin Wu},
    publisher={O'Reilly Media},
    note={\url{https://www.amazon.com/Deep-Learning-Life-Sciences-Microscopy/dp/1492039837}},
    year={2019}
}

@article{riniker_open-source_2013,
    title = {Open-source platform to benchmark fingerprints for ligand-based virtual screening},
    volume = {5},
    issn = {1758-2946},
    url = {https://doi.org/10.1186/1758-2946-5-26},
    doi = {10.1186/1758-2946-5-26},
    number = {1},
    urldate = {2025-10-15},
    journal = {Journal of Cheminformatics},
    author = {Riniker, Sereina and Landrum, Gregory A.},
    month = may,
    year = {2013},
    pages = {26},
}

@article{orosz_comparison_2022,
    title = {Comparison of {Descriptor}- and {Fingerprint} {Sets} in {Machine} {Learning} {Models} for {ADME}-{Tox} {Targets}},
    volume = {10},
    issn = {2296-2646},
    url = {https://www.frontiersin.org/journals/chemistry/articles/10.3389/fchem.2022.852893/full},
    doi = {10.3389/fchem.2022.852893},
    language = {English},
    urldate = {2025-10-15},
    journal = {Frontiers in Chemistry},
    author = {Orosz, Almos and Héberger, Károly and Rácz, Anita},
    month = jun,
    year = {2022},
    note = {Publisher: Frontiers},
}

@article{schimunek2025mhnfs,
  title={MHNfs: Prompting In-Context Bioactivity Predictions for Low-Data Drug Discovery},
  author={Schimunek, Johannes and Luukkonen, Sohvi and Klambauer, Günter},
  journal={Journal of Chemical Information and Modeling},
  volume={65},
  number={9},
  pages={4243--4250},
  year={2025},
  publisher={ACS Publications}
}

@inproceedings{chenmeta,
  title={Meta-learning Adaptive Deep Kernel Gaussian Processes for Molecular Property Prediction},
  author={Chen, Wenlin and Tripp, Austin and Hern{\'a}ndez-Lobato, Jos{\'e} Miguel},
  year={2023},
  booktitle={The Eleventh International Conference on Learning Representations}
}
\bibliographystyle{apalike}
\clearpage

\appendix
\counterwithin{figure}{section}
\counterwithin{table}{section}
\renewcommand\thefigure{\thesection\arabic{figure}}
\renewcommand\thetable{\thesection\arabic{table}}
\renewcommand{\contentsname}{Appendix}

\resumetocwriting

\tableofcontents

\newpage

\section{Details on the Leaderboard}

\subsection{User Request to Add a Model}
To integrate a new model into the leaderboard, users submit an evaluation request through the leaderboard frontend using a standardized submission template. Each submission must include a model card following the guidelines of \citet{mitchell2019model}, along with metadata specifying the model’s Hugging~Face Space URL and corresponding Git commit hash:
\begin{center}
\begin{tcolorbox}[enhanced, breakable, width=0.9\textwidth, title=Model card]
\small

\textbf{Model details}
\begin{itemize}
    \item Model name:
    \item Developer:
    \item Paper URL:
    \item Model type / architecture:
    \item Inference:
    \item Model version:
    \item Model date:
    \item Reproducibility: Code for full training is available and enables reproducible retraining of the model from scratch.
\end{itemize}

\medskip
\textbf{Intended use:}\par
\vspace{1.em}

\textbf{Metric:}\par
\vspace{1.em}

\textbf{Training data:}\par
\vspace{1.em}

\textbf{Evaluation data:}\par
\vspace{1.em}

\textbf{Hugging Face Space:}\par
\vspace{1.em}

\end{tcolorbox}
\end{center}

The Hugging~Face Space repository must include all code necessary to reproduce the submitted model from scratch. 

Submitted models must provide predictions for all 647 test molecules and all twelve Tox21 targets. The leaderboard validates that predictions for every molecule–target pair are present and raises an error otherwise. This ensures a fair comparison between models and prevents potential test score hacking by excluding difficult molecules.

\subsection{Submission verification step}
\label{appsec:submission_verification}
The Tox21 leaderboard emphasizes reproducibility. To ensure this, all submissions are manually reviewed before being displayed. Users first submit an evaluation request and provide all required information. The leaderboard automatically retrieves and evaluates the submitted model's predictions, and subsequently marks the obtained performance results as preliminary. After a manual review — verifying (a) completeness of the provided information and (b) reproducibility of model training 
-- the evaluation and its results are marked as approved and subsequently published on the leaderboard.

\newpage
\section{Details on Tox21}
Table~\ref{tab:targets} lists the twelve targets included in the Tox21 dataset, together with short descriptions.
\begin{table}[H]
\small
\centering
\caption{Tox21 assay targets and descriptions}
\label{tab:targets}
\begin{tabular}{ll}
\hline
\textbf{Target} & \textbf{Description} \\
\hline
NR-AR & Androgen Receptor - involved in male hormone signaling \\
NR-AR-LBD & Androgen Receptor Ligand Binding Domain - direct binding to androgen receptor \\
NR-AhR & Aryl Hydrocarbon Receptor - responds to environmental chemicals \\
NR-Aromatase & Aromatase enzyme - converts androgens to estrogens \\
NR-ER & Estrogen Receptor - involved in female hormone signaling \\
NR-ER-LBD & Estrogen Receptor Ligand Binding Domain - direct binding to estrogen receptor \\
NR-PPAR-gamma & Peroxisome Proliferator-Activated Receptor Gamma - regulates metabolism \\
SR-ARE & Antioxidant Response Element - responds to oxidative stress \\
SR-ATAD5 & ATAD5 - involved in DNA replication and genome stability \\
SR-HSE & Heat Shock Response Element - responds to cellular stress \\
SR-MMP & Mitochondrial Membrane Potential - indicates mitochondrial function \\
SR-p53 & p53 tumor suppressor - activated by DNA damage and stress \\
\hline
\end{tabular}
\end{table}

Table~\ref{tab:tox21_splits} compares the original Tox21 dataset with the 
preprocessed versions distributed through MoleculeNet and DeepChem. The 
redistributed data contains fewer samples overall and exhibits substantial 
differences in train/validation/test splits, percentages of missing values, 
and active compound ratios compared to the original dataset.

\begin{table}[H]
\centering
\caption{Comprehensive comparison of Tox21 dataset splits. Original refers to the data as originally released in the challenge, while DC denotes DeepChem \citep{Ramsundar-et-al-2019} splits (Random, Scaffold, and Stratified).}
\label{tab:tox21_splits}
\small
\begin{tabular}{@{}llrrrrr@{}}
\toprule
\textbf{Split} & \textbf{Dataset} & \textbf{Total} & \textbf{Unique} & \textbf{Labeled \%} & \textbf{Missing \%} & \textbf{Active \%} \\
\midrule
\multirow{4}{*}{Train} 
& Original        & 11,764 & 8,043 & 69.7 & 30.3 & 7.3 \\
& DC-Random       & 6,258  & 6,258 & 82.9 & 17.1 & 7.4 \\
& DC-Scaffold     & 6,258  & 6,258 & 84.7 & 15.3 & 7.1 \\
& DC-Stratified   & 6,258  & 6,258 & 82.8 & 17.2 & 7.5 \\
\midrule
\multirow{4}{*}{Validation} 
& Original        & 296    & 296   & 88.1 & 11.9 & 8.2 \\
& DC-Random       & 782    & 782   & 82.5 & 17.5 & 8.2 \\
& DC-Scaffold     & 782    & 782   & 76.1 & 23.9 & 9.7 \\
& DC-Stratified   & 782    & 782   & 82.1 & 17.9 & 7.6 \\
\midrule
\multirow{4}{*}{Test} 
& Original        & 647    & 645   & 89.8 & 10.2 & 7.0 \\
& DC-Random       & 783    & 783   & 84.0 & 16.0 & 7.5 \\
& DC-Scaffold     & 783    & 783   & 76.1 & 23.9 & 9.3 \\
& DC-Stratified   & 783    & 783   & 84.6 & 15.4 & 7.4 \\
\bottomrule
\end{tabular}
\end{table}


To further illustrate these discrepancies, Table~\ref{tab:tox21_all_assays} 
provides a detailed breakdown of data availability and class balance for each 
assay across all dataset variants and splits. The variation in missing value 
percentages and activity ratios across splits highlights potential issues with 
data preprocessing and splitting strategies in the redistributed versions.

\begin{table}[htbp]
\centering
\caption{Comparison of label availability, missing data, and class balance across 
train/validation/test splits for all Tox21 assays. Statistics shown for the 
original dataset and three DeepChem splitting strategies (Random, Scaffold, 
Stratified). Lab.\% = labeled samples, Miss.\% = missing values, Act.\% = active 
compounds.}
\label{tab:tox21_all_assays}
\scriptsize
\begin{tabular}{@{}llccccccccc@{}}
\toprule
\multirow{2}{*}{\textbf{Assay}} & \multirow{2}{*}{\textbf{Dataset}} & \multicolumn{3}{c}{\textbf{Train}} & \multicolumn{3}{c}{\textbf{Validation}} & \multicolumn{3}{c}{\textbf{Test}} \\
\cmidrule(lr){3-5} \cmidrule(lr){6-8} \cmidrule(lr){9-11}
& & Lab.\% & Miss.\% & Act.\% & Lab.\% & Miss.\% & Act.\% & Lab.\% & Miss.\% & Act.\% \\
\midrule
\multirow{4}{*}{SR-ARE} 
& Original      & 60.9 & 39.1 & 15.3 & 79.1 & 20.9 & 20.5 & 85.8 & 14.2 & 16.8 \\
& DC-Random     & 74.4 & 25.6 & 16.1 & 73.4 & 26.6 & 15.2 & 75.9 & 24.1 & 17.5 \\
& DC-Scaffold   & 77.8 & 22.2 & 14.7 & 60.5 & 39.5 & 22.4 & 61.4 & 38.6 & 24.5 \\
& DC-Stratified & 74.7 & 25.3 & 16.1 & 71.1 & 28.9 & 16.9 & 75.5 & 24.5 & 15.9 \\
\cmidrule(lr){1-11}
\multirow{4}{*}{SR-HSE} 
& Original      & 69.3 & 30.7 & 5.3 & 90.2 & 9.8 & 3.7 & 94.3 & 5.7 & 3.6 \\
& DC-Random     & 82.1 & 17.9 & 5.5 & 83.1 & 16.9 & 7.4 & 85.6 & 14.4 & 6.3 \\
& DC-Scaffold   & 84.9 & 15.1 & 5.3 & 73.4 & 26.6 & 7.7 & 73.1 & 26.9 & 8.2 \\
& DC-Stratified & 82.6 & 17.4 & 5.8 & 81.8 & 18.2 & 5.8 & 83.4 & 16.6 & 5.7 \\
\cmidrule(lr){1-11}
\multirow{4}{*}{SR-MMP} 
& Original      & 62.2 & 37.8 & 15.6 & 80.4 & 19.6 & 16.0 & 83.9 & 16.1 & 11.0 \\
& DC-Random     & 74.5 & 25.5 & 15.8 & 72.6 & 27.4 & 16.0 & 73.1 & 26.9 & 15.9 \\
& DC-Scaffold   & 76.3 & 23.7 & 14.9 & 64.9 & 35.1 & 21.9 & 66.4 & 33.6 & 18.5 \\
& DC-Stratified & 73.7 & 26.3 & 15.9 & 74.8 & 25.2 & 15.7 & 77.5 & 22.5 & 15.2 \\
\cmidrule(lr){1-11}
\multirow{4}{*}{SR-p53} 
& Original      & 73.4 & 26.6 & 6.2 & 90.9 & 9.1 & 10.4 & 95.2 & 4.8 & 6.7 \\
& DC-Random     & 86.3 & 13.7 & 6.1 & 85.9 & 14.1 & 7.6 & 88.8 & 11.2 & 5.9 \\
& DC-Scaffold   & 87.7 & 12.3 & 5.0 & 82.9 & 17.1 & 11.6 & 80.5 & 19.5 & 11.4 \\
& DC-Stratified & 86.3 & 13.7 & 6.3 & 85.6 & 14.4 & 6.3 & 88.6 & 11.4 & 6.1 \\
\cmidrule(lr){1-11}
\multirow{4}{*}{SR-ATAD5} 
& Original      & 77.3 & 22.7 & 3.7 & 91.9 & 8.1 & 9.2 & 96.1 & 3.9 & 6.1 \\
& DC-Random     & 90.2 & 9.8 & 3.5 & 91.2 & 8.8 & 5.3 & 90.5 & 9.5 & 4.2 \\
& DC-Scaffold   & 91.5 & 8.5 & 3.4 & 84.9 & 15.1 & 5.3 & 85.8 & 14.2 & 4.9 \\
& DC-Stratified & 90.2 & 9.8 & 3.8 & 89.5 & 10.5 & 3.7 & 91.7 & 8.3 & 3.6 \\
\midrule
\multirow{4}{*}{NR-AR} 
& Original      & 79.6 & 20.4 & 4.1 & 98.6 & 1.4 & 1.0 & 90.6 & 9.4 & 2.0 \\
& DC-Random     & 92.6 & 7.4 & 4.3 & 93.6 & 6.4 & 5.1 & 93.5 & 6.5 & 3.0 \\
& DC-Scaffold   & 93.0 & 7.0 & 4.3 & 92.5 & 7.5 & 4.3 & 91.3 & 8.7 & 3.8 \\
& DC-Stratified & 92.6 & 7.4 & 4.2 & 92.8 & 7.2 & 4.3 & 93.7 & 6.3 & 4.2 \\
\cmidrule(lr){1-11}
\multirow{4}{*}{NR-AR-LBD} 
& Original      & 73.1 & 26.9 & 3.5 & 85.5 & 14.5 & 1.6 & 90.0 & 10.0 & 1.4 \\
& DC-Random     & 86.4 & 13.6 & 3.5 & 85.0 & 15.0 & 4.8 & 86.6 & 13.4 & 2.4 \\
& DC-Scaffold   & 87.9 & 12.1 & 3.5 & 80.2 & 19.8 & 4.0 & 79.7 & 20.3 & 3.0 \\
& DC-Stratified & 86.1 & 13.9 & 3.5 & 85.3 & 14.7 & 3.6 & 88.8 & 11.2 & 3.6 \\
\cmidrule(lr){1-11}
\multirow{4}{*}{NR-AhR} 
& Original      & 69.4 & 30.6 & 11.6 & 91.9 & 8.1 & 11.4 & 94.3 & 5.7 & 12.0 \\
& DC-Random     & 83.6 & 16.4 & 11.6 & 83.6 & 16.4 & 11.9 & 84.0 & 16.0 & 12.6 \\
& DC-Scaffold   & 84.6 & 15.4 & 11.1 & 79.2 & 20.8 & 14.1 & 80.3 & 19.7 & 14.6 \\
& DC-Stratified & 83.5 & 16.5 & 11.8 & 83.1 & 16.9 & 11.8 & 85.3 & 14.7 & 11.5 \\
\cmidrule(lr){1-11}
\multirow{4}{*}{NR-Aromatase} 
& Original      & 61.4 & 38.6 & 5.0 & 72.3 & 27.7 & 8.4 & 81.6 & 18.4 & 7.4 \\
& DC-Random     & 74.5 & 25.5 & 5.2 & 73.4 & 26.6 & 4.9 & 74.3 & 25.7 & 5.3 \\
& DC-Scaffold   & 76.4 & 23.6 & 4.3 & 65.1 & 34.9 & 8.8 & 66.8 & 33.2 & 9.0 \\
& DC-Stratified & 74.1 & 25.9 & 5.2 & 74.2 & 25.8 & 5.2 & 76.4 & 23.6 & 5.0 \\
\cmidrule(lr){1-11}
\multirow{4}{*}{NR-ER} 
& Original      & 65.4 & 34.6 & 12.2 & 89.5 & 10.5 & 10.2 & 79.7 & 20.3 & 9.9 \\
& DC-Random     & 78.8 & 21.2 & 12.7 & 78.5 & 21.5 & 12.2 & 82.1 & 17.9 & 13.8 \\
& DC-Scaffold   & 81.1 & 18.9 & 12.7 & 70.8 & 29.2 & 13.5 & 70.8 & 29.2 & 12.6 \\
& DC-Stratified & 79.2 & 20.8 & 12.8 & 77.5 & 22.5 & 13.0 & 79.3 & 20.7 & 12.7 \\
\cmidrule(lr){1-11}
\multirow{4}{*}{NR-ER-LBD} 
& Original      & 74.4 & 25.6 & 5.1 & 97.0 & 3.0 & 3.5 & 92.7 & 7.3 & 3.3 \\
& DC-Random     & 88.7 & 11.3 & 5.0 & 88.6 & 11.4 & 6.1 & 90.2 & 9.8 & 4.4 \\
& DC-Scaffold   & 90.0 & 10.0 & 5.3 & 84.9 & 15.1 & 4.4 & 83.4 & 16.6 & 3.2 \\
& DC-Stratified & 88.7 & 11.3 & 5.0 & 87.2 & 12.8 & 5.1 & 91.2 & 8.8 & 4.9 \\
\cmidrule(lr){1-11}
\multirow{4}{*}{NR-PPAR-gamma} 
& Original      & 69.6 & 30.4 & 2.7 & 90.2 & 9.8 & 5.6 & 93.5 & 6.5 & 5.1 \\
& DC-Random     & 82.4 & 17.6 & 2.8 & 80.9 & 19.1 & 4.6 & 83.5 & 16.5 & 2.3 \\
& DC-Scaffold   & 84.6 & 15.4 & 2.5 & 73.6 & 26.4 & 5.6 & 73.4 & 26.6 & 3.8 \\
& DC-Stratified & 82.2 & 17.8 & 2.9 & 82.1 & 17.9 & 3.0 & 83.9 & 16.1 & 2.9 \\
\bottomrule
\end{tabular}
\end{table}

\newpage
\section{Details on the Baselines}\label{appsec:baselines}

\subsection{Molecule Features}

Feature-based bioactivity models are sensitive to the choice of molecular representation \citep{riniker_open-source_2013,orosz_comparison_2022}. To isolate architectural effects, we fixed the input features across all feature-based baselines (RF, XGBoost, SNN, TabPFN) and used the same 9,385 molecular features for every model. The final input vector was a concatenation of: 
\begin{itemize}
    \item 8192-bit folded ECFP6 count fingerprints,
    \item 166 MACCS keys,
    \item 200 selected RDKit molecular descriptors, and
    \item 827 descriptors associated with known toxicity patterns.
\end{itemize}

The selection of RDKit descriptors was fixed across models and is specified in the Hugging Face Space for each feature-based baseline. Before modeling, we optionally applied low-variance and high-correlation feature filtering, quantization of RDKit descriptors, and feature normalization. The corresponding respective variance and correlation thresholds, along with flags enabling quantilization and normalization, were treated as hyperparameters to be optimized for each model. For TabPFN, we only used the 500 highest-variance features after filtering, as TabPFN-v2 supports at most 500 features.

\subsection{Self-Normalizing Neural Network}
\begin{center}
\begin{tcolorbox}[enhanced, breakable, width=0.9\textwidth, title=Model card -- SNN]
\small

\textbf{Model details}
\begin{itemize}
    \item \textbf{Model name:} Self-Normalizing Neural Network Tox21 Baseline
    \item \textbf{Developer:} JKU Linz
    \item \textbf{Paper URL:} \url{https://proceedings.neurips.cc/paper_files/paper/2017/hash/5d44ee6f2c3f71b73125876103c8f6c4-Abstract.html}
    \item \textbf{Model type / architecture:}
    \begin{itemize}
        \item Self-Normalizing Neural Network implemented using \texttt{PyTorch}.
        \item \textbf{Hyperparameters:} \url{https://huggingface.co/spaces/ml-jku/tox21_snn_classifier/blob/main/config/config.json}
        \item A multitask network is trained for all Tox21 targets.
    \end{itemize}
    \item \textbf{Inference:} Access via FastAPI endpoint. Upon receiving a Tox21 prediction request, the model generates and returns predictions for all Tox21 targets simultaneously.
    \item \textbf{Model version:} v0
    \item \textbf{Model date:} 14.10.2025
    \item \textbf{Reproducibility:} Code for full training is available and enables reproducible retraining of the model from scratch.
\end{itemize}

\medskip
\textbf{Intended use:}   \\
This model serves as a baseline benchmark for evaluating and comparing toxicity prediction methods across the 12 pathway assays of the Tox21 dataset. It is not intended for clinical decision-making without experimental validation.

\medskip
\textbf{Metric:}   \\
Each Tox21 task is evaluated using the area under the receiver operating characteristic curve (AUC). Overall performance is reported as the mean AUC across all individual tasks.

\medskip
\textbf{Training data:} \\
Tox21 training and validation sets.

\textbf{Evaluation data:}  \\
Tox21 test set.
\newpage
\medskip
\textbf{Hugging Face Space:} \\
\url{https://huggingface.co/spaces/ml-jku/tox21_snn_classifier/blob/main/README.md}

\end{tcolorbox}
\end{center}


\subsection{Random Forest}
\begin{center}
\begin{tcolorbox}[enhanced,breakable,width=0.9\textwidth,title=Model card -- RF]
\small


\textbf{Model details}
\begin{itemize}
    \item \textbf{Model name:} Random Forest Tox21 Baseline
    \item \textbf{Developer:} JKU Linz
    \item \textbf{Paper URL:} \url{https://link.springer.com/article/10.1023/A:1010933404324}
    \item \textbf{Model type / architecture:}
    \begin{itemize}
        \item Random Forest implemented using \texttt{sklearn.RandomForestClassifier}.
        \item \textbf{Hyperparameters:} \url{https://huggingface.co/spaces/ml-jku/tox21_rf_classifier/blob/main/config/config.json}
        \item A separate single-task RF is trained for each Tox21 target.
    \end{itemize}
    \item \textbf{Inference:} Access via FastAPI. Upon a Tox21 prediction request, a target-specific RF model is called separately for each target; outputs are collected across all single-task models and returned.
    \item \textbf{Model version:} v0
    \item \textbf{Model date:} 14.10.2025
    \item \textbf{Reproducibility:} Code for full training is available and enables retraining from scratch.
\end{itemize}

\medskip
\textbf{Intended use:} \\
This model serves as a baseline for evaluating and comparing toxicity prediction methods across the 12 Tox21 pathway assays. It is not intended for clinical decision-making without experimental validation.

\medskip
\textbf{Metric:} \\
Each Tox21 task is evaluated using the area under the receiver operating characteristic curve (AUC). Overall performance is reported as the mean AUC across all tasks.

\medskip
\textbf{Training data:} \\
Tox21 training and validation sets.

\textbf{Evaluation data:} \\
Tox21 test set.

\medskip
\textbf{Hugging~Face Space:} \\
\url{https://huggingface.co/spaces/ml-jku/tox21_rf_classifier/blob/main/README.md}

\end{tcolorbox}
\end{center}

\subsection{Chemprop}
\begin{center}
\begin{tcolorbox}[enhanced, breakable, width=0.9\textwidth, title=Model card -- Chemprop]
\small

\textbf{Model details}
\begin{itemize}
    \item \textbf{Model name:} Chemprop Message Passing Neural Network Tox21 Baseline
    \item \textbf{Developer:} MIT (trained by JKU Linz)
    \item \textbf{Paper URL:} \url{https://pubs.acs.org/doi/full/10.1021/acs.jcim.3c01250}
    \item \textbf{Model type / architecture:}
    \begin{itemize}
        \item Message Passing Neural Network implemented using \texttt{Chemprop}.
        \item \textbf{Hyperparameters:} \url{https://huggingface.co/spaces/ml-jku/tox21_chemprop_classifier/blob/main/config/config.toml}
        \item A multitask network is trained for all Tox21 targets.
    \end{itemize}
    \item \textbf{Inference:} Access via FastAPI endpoint. Upon receiving a Tox21 prediction request, the model generates and returns predictions for all Tox21 targets simultaneously.
    \item \textbf{Model version:} v0
    \item \textbf{Model date:} 14.10.2025
    \item \textbf{Reproducibility:} Code for full training is available and enables reproducible retraining of the model from scratch.
\end{itemize}

\medskip
\textbf{Intended use:}   \\
This model serves as a baseline benchmark for evaluating and comparing toxicity prediction methods across the 12 pathway assays of the Tox21 dataset. It is not intended for clinical decision-making without experimental validation.

\medskip
\textbf{Metric:}   \\
Each Tox21 task is evaluated using the area under the receiver operating characteristic curve (AUC). Overall performance is reported as the mean AUC across all individual tasks.

\medskip
\textbf{Training data:}  \\
Tox21 training and validation sets.

\textbf{Evaluation data:}  \\
Tox21 test set.

\medskip
\textbf{Hugging Face Space:} \\
\url{https://huggingface.co/spaces/ml-jku/tox21_chemprop_classifier/blob/main/README.md}

\end{tcolorbox}
\end{center}


\subsection{XGBoost}
\begin{center}
\begin{tcolorbox}[enhanced, breakable, width=0.9\textwidth, title=Model card -- XGBoost]
\small

\textbf{Model details}
\begin{itemize}
    \item \textbf{Model name:} XGBoost Tox21 Baseline
    \item \textbf{Developer:} JKU Linz
    \item \textbf{Paper URL:} \url{https://cran.ms.unimelb.edu.au/web/packages/xgboost/vignettes/xgboost.pdf}
    \item \textbf{Model type / architecture:}
    \begin{itemize}
        \item Extreme Gradient Boosting implemented using \texttt{xgboost}.
        \item \textbf{Hyperparameters:} \url{https://huggingface.co/spaces/ml-jku/tox21_xgboost_classifier/blob/main/config/config.json}
        \item A separate single-task RF is trained for each Tox21 target.
    \end{itemize}
    \item \textbf{Inference:} Access via FastAPI endpoint. Upon receiving a Tox21 prediction request, the model generates and returns predictions for all Tox21 targets simultaneously.
    \item \textbf{Model version:} v0
    \item \textbf{Model date:} 14.10.2025
    \item \textbf{Reproducibility:} Code for full training is available and enables reproducible retraining of the model from scratch.
\end{itemize}

\medskip
\textbf{Intended use:}   \\
This model serves as a baseline benchmark for evaluating and comparing toxicity prediction methods across the 12 pathway assays of the Tox21 dataset. It is not intended for clinical decision-making without experimental validation.

\medskip
\textbf{Metric:}   \\
Each Tox21 task is evaluated using the area under the receiver operating characteristic curve (AUC). Overall performance is reported as the mean AUC across all individual tasks.

\medskip
\textbf{Training data:}  \\
Tox21 training and validation sets.

\textbf{Evaluation data:} \\
Tox21 test set.

\medskip
\textbf{Hugging Face Space:} \\
\url{https://huggingface.co/spaces/ml-jku/tox21_xgboost_classifier/blob/main/README.md}

\end{tcolorbox}
\end{center}



\subsection{Graph Isomorphism Network}
\begin{center}
\begin{tcolorbox}[enhanced,breakable,width=0.9\textwidth,title=Model card -- GIN]
\small

\textbf{Model details}
\begin{itemize}
    \item \textbf{Model name:} Graph Isomorphism Network Tox21 Baseline
    \item \textbf{Developer:} MIT \& Stanford (trained by JKU Linz)
    \item \textbf{Paper URL:} \url{https://arxiv.org/abs/1810.00826}
    \item \textbf{Model type / architecture:}
    \begin{itemize}
        \item Graph Isomorphism Network implemented using \texttt{PyTorch}.
        \item \textbf{Hyperparameters:} \url{https://huggingface.co/spaces/ml-jku/tox21_gin_classifier/blob/main/config/config.json}
        \item A multitask network is trained for all Tox21 targets.
    \end{itemize}
    \item \textbf{Inference:} Access via FastAPI endpoint. Upon a Tox21 prediction request, the model generates and returns predictions for all Tox21 targets simultaneously.
    \item \textbf{Model version:} v0
    \item \textbf{Model date:} 14.10.2025
    \item \textbf{Reproducibility:} Code for full training is available and enables retraining from scratch.
\end{itemize}

\medskip
\textbf{Intended use:} \\
This model serves as a baseline benchmark for evaluating and comparing toxicity prediction methods across the 12 pathway assays of the Tox21 dataset. It is not intended for clinical decision-making without experimental validation.

\medskip
\textbf{Metric:} \\
Each Tox21 task is evaluated using the area under the receiver operating characteristic curve (AUC). Overall performance is reported as the mean AUC across all individual tasks.

\medskip
\textbf{Training data:} \\
Tox21 training and validation sets.

\textbf{Evaluation data:} \\
Tox21 test set.

\medskip
\textbf{Hugging Face Space:}\\
\url{https://huggingface.co/spaces/ml-jku/tox21_gin_classifier/blob/main/README.md}

\end{tcolorbox}
\end{center}


\subsection{TabPFN}
\begin{center}
\begin{tcolorbox}
[enhanced,breakable,width=0.9\textwidth,title=Model card -- TabPFN]
\small

\textbf{Model details}
\begin{itemize}
    \item \textbf{Model name:} TabPFN Tox21 Baseline
    \item \textbf{Developer:} PriorLabs (trained by JKU Linz)
    \item \textbf{Paper URL:} \url{https://openreview.net/forum?id=eu9fVjVasr4}
    \item \textbf{Model type / architecture:}
    \begin{itemize}
        \item TabPFN implemented using \texttt{tabpfn.TabPFNClassifier}.
        \item A separate single-task TabPFN model is trained for each Tox21 target.
    \end{itemize}
    \item \textbf{Inference:} Local inference. Provide SMILES inputs to \texttt{predict()} in \texttt{predict.py} after completing setup and installation. A separate TabPFN model is called for each target. Outputs are collected across all single-task models and returned.
    \item \textbf{Model version:} v0
    \item \textbf{Model date:} 12.10.2025
    \item \textbf{Reproducibility:} Code for full training is available and enables refitting of TabPFN from scratch.
    \item \textbf{License / availability}: Open-weight model available under Prior Labs License
\end{itemize}

\medskip
\textbf{Intended use:} \\
This model serves as a baseline benchmark for evaluating and comparing toxicity prediction methods across the 12 pathway assays of the Tox21 dataset. It is not intended for clinical decision-making without experimental validation.

\medskip
\textbf{Metric:} \\
Each Tox21 task is evaluated using the area under the receiver operating characteristic curve (AUC). Overall performance is reported as the mean AUC across all individual tasks.

\medskip
\textbf{Training data:} \\
Tox21 training and validation sets.

\textbf{Evaluation data:} \\
Tox21 test set.

\medskip
\textbf{Hugging Face Space:}\\
\url{https://huggingface.co/spaces/ml-jku/tox21_tabpfn_classifier/blob/main/README.md}

\end{tcolorbox}
\end{center}


\subsection{GPT-OSS}
\begin{center}
\begin{tcolorbox}[enhanced,breakable,width=0.9\textwidth,title=Model card -- GPT-OSS]
\small

\textbf{Model details}
\begin{itemize}
    \item \textbf{Model name:} GPT-OSS 120B
    \item \textbf{Developer:} OpenAI
    \item \textbf{Model card URL:} \url{https://openai.com/index/gpt-oss-model-card}
    \item \textbf{Model type / architecture:}
    \begin{itemize}
        \item Mixture-of-Experts (MoE) Transformer architecture
        \item 36 transformer layers, 117B total parameters (5.1B active per forward pass)
        \item 128 experts per MoE layer with top-4 expert routing
        \item Hidden dimension 2880, 64 attention heads, Grouped Query Attention (group size 8)
        \item Context length 128K tokens
    \end{itemize}
    \item \textbf{Inference:}
    \begin{itemize}
        \item Context length 131{,}072 tokens
        \item Temperature 0.7
        \item Reasoning level: \texttt{high}
        \item Prompting strategy: see \ref{app:gpt_oss_prompt_config}
        \item Inference engine: vLLM
        \item Rollout strategy: five independent rollouts per molecule with aggregated predictions
    \end{itemize}
    \item \textbf{Model version:} gpt-oss-120b
    \item \textbf{Model date:} August 5, 2025
    \item \textbf{License / availability:} Open-weight model available under the Apache 2.0 license; can be fine-tuned for specialized use cases.
\end{itemize}

\medskip
\textbf{Intended use:} \\
This open-weight model is designed for reasoning, agentic, and developer tasks. It is intended for research and development purposes, including code generation, technical analysis, scientific reasoning, and complex problem solving.

\medskip
\textbf{Metric:} \\
Each Tox21 task is evaluated using the area under the receiver-operating-characteristic curve (AUC). Overall performance is reported as the mean AUC across all tasks.

\medskip
\textbf{Training data:} \\
Proprietary dataset (composition not publicly disclosed). Evaluated without additional fine-tuning or few-shot prompting.

\textbf{Evaluation data:} \\
Tox21 test set.

\medskip
\textbf{Hugging Face Model Repository}
\begin{itemize}
    \item \textbf{Repository:} \url{https://huggingface.co/openai/gpt-oss-120b}
    \item \textbf{Inference engine:} vLLM \citep{kwon2023vllm}
\end{itemize}

\end{tcolorbox}
\end{center}

\subsubsection*{Prompt Template \& Model Configuration}
\label{app:gpt_oss_prompt_config}
The language model was queried using a two-part prompt structure with system and user roles:

\begin{center}
\begin{tcolorbox}[enhanced,breakable,width=0.9\textwidth,title=Prompt structure]
\small

\textbf{System:} You are an expert in molecular toxicity prediction.  
Analyze molecules and provide probability scores between \texttt{0.000} and \texttt{1.000}.  
Always respond with up to three decimal places.

\medskip
\textbf{User:} Analyze whether this molecule is likely to be toxic in the \texttt{\{target\}} assay.

\smallskip
\textbf{Target:} \texttt{\{description\}} \\
\textbf{SMILES:} \texttt{\{smiles\}}

\medskip
Provide only a probability between \texttt{0.000} and \texttt{1.000} indicating the likelihood of toxicity.  
Respond with only the number.
\end{tcolorbox}
\end{center}

The template variables \texttt{\{target\}}, \texttt{\{description\}}, and \texttt{\{smiles\}} were populated for each prediction task. Table~\ref{tab:targets} lists all twelve Tox21 assay targets and their corresponding descriptions used in the prompts.

\textbf{Model Configuration:}
 We used the GPT-OSS 120B model \citep{openai2025gptoss120bgptoss20bmodel} with default sampling settings as suggested by the model developers. This means a context size of 131,072 tokens, a sampling temperature of 0.7, and a reasoning setting of \texttt{high}. We perform five individual rollouts using vLLM \citep{kwon2023vllm} as our inference engine.

\section{Details on the FastAPI Template}

\textbf{Overview.}
We provide a minimal FastAPI template that exposes a single batch inference endpoint, \verb|/predict|.  
The endpoint accepts a list of SMILES strings and returns a nested dictionary of prediction scores, keyed first by input SMILES and then by assay target.
The template is built in a modular way so that users can easily adapt it for their models.

\textbf{FastAPI server.}
The template mounts \verb|/predict| and delegates to a pure function \verb|predict_fn|. Code is stored in \verb|app.py| and does not need to be adapted by the users:

\begin{verbatim}
from fastapi import FastAPI
from pydantic import BaseModel
from typing import Dict, List

app = FastAPI()

class Request(BaseModel):
    smiles: List[str]

class Response(BaseModel):
    predictions: Dict[str, Dict[str, float]]
    model_info: Dict[str, str]

@app.post("/predict", response_model=Response)
def predict(request: Request):
    predictions = predict_fn(request.smiles)
    return {
        "predictions": predictions,
        "model_info": {"name": "tox21_model", "version": "1.0.0"},
    }
\end{verbatim}

\textbf{Toxicity prediction wrapper.}
The function \verb|predict_fn| - stored in \verb|predict.py| - encapsulates preprocessing and inference.  
Users should only replace the data preprocessing and model code, keeping the input/output contract unchanged:

\begin{verbatim}
# Pseudo code; for a working example see one of the baseline models
def predict_fn(smiles_list: list[str]) -> dict[str, dict[str, float]]:
    """Return {'<smiles>': {'<target>': <pred>}} for each input SMILES.
    """

    # --- Preprocessing (replace with your pipeline) ---
    scaler = load_pickle("assets/scaler.pkl")
    features = create_features(smiles_list, scaler=scaler)

    # --- Model (replace with your model) ---
    model = Model(seed=42)
    model.load_model("assets/trained_model")

    # --- Inference ---
    predictions = model(features)

    # --- Correct output format ---
    for smiles in smiles_list:
        for target in targets:
            output_dict[smiles][target] = predictions[...]

    return output_dict
\end{verbatim}

\textbf{Interface guarantees and validation.}
For leaderboard compatibility, the service must:
\begin{itemize}
    \item Accept variable batch sizes (the orchestrator may split requests).
    \item Return predictions for \emph{every} SMILES--target pair. Missing entries or NaN values cause the submission to be rejected.
    \item Be deterministic given fixed weights and preprocessing artifacts (set random seeds if necessary).
\end{itemize}

\textbf{Example request:}
\begin{verbatim}
curl -X POST https://ml-jku-tox21-gin-classifier.hf.space/predict \
  -H "Content-Type: application/json" \
  -d '{"smiles": ["CCO", "c1ccccc1"]}'
\end{verbatim}

\textbf{Returned JSON:}
\begin{verbatim}
{
  "predictions": {"CCO":{"NR-AhR":0.005747087765485048,
                         "NR-AR":0.001738760736770928,
                         "NR-AR-LBD":0.00021425147133413702,
                         ...
                         "SR-p53":0.0007309493375942111},
                  "c1ccccc1":{
                         ...
                  }},‚
  "model_info": {"name": "Tox21 GIN classifier", "version": "1.0.0"}
}
\end{verbatim}

\end{document}